\documentclass[10pt,twocolumn,letterpaper]{article}

\usepackage[pagenumbers]{cvpr} 

\definecolor{cvprblue}{rgb}{0.21,0.49,0.74}
\usepackage[pagebackref,breaklinks,colorlinks,allcolors=cvprblue]{hyperref}

\def\paperID{3936} 
\def\confName{CVPR}
\def\confYear{2026}

\usepackage{xspace} 
\usepackage{graphicx}
\usepackage{multirow}
\usepackage{pifont} 
\usepackage{amsmath}
\usepackage{amssymb}
\usepackage{subcaption}
\newcommand{\cmark}{\ding{51}}
\newcommand{\xmark}{\ding{55}}

\newcommand{\method}{OpenDance\xspace}
\newcommand{\dataset}{OpenDanceSet\xspace}
\newcommand{\model}{OpenDanceNet\xspace}

\title{\method: Multimodal Controllable 3D Dance Generation with Large-scale Internet Data}

\author{
    Jinlu Zhang\textsuperscript{1}\thanks{Equal contribution.} \quad
    Zixi Kang\textsuperscript{1}\footnotemark[1] \quad
    Libin Liu\textsuperscript{1} \quad
    Jianlong Chang\textsuperscript{2} \quad
    Qi Tian\textsuperscript{2} \quad
    Feng Gao\textsuperscript{1} \quad
    Yizhou Wang\textsuperscript{1} \\[2mm]
    \textsuperscript{1}Peking University \qquad  
    \textsuperscript{2}Huawei Cloud \\
}

\begin{document}
\twocolumn[{
\renewcommand\twocolumn[1][]{#1}
\maketitle
\begin{center}
    \centering
    \vspace{-10pt}
    \captionsetup{type=figure}
    \includegraphics[width=\linewidth]{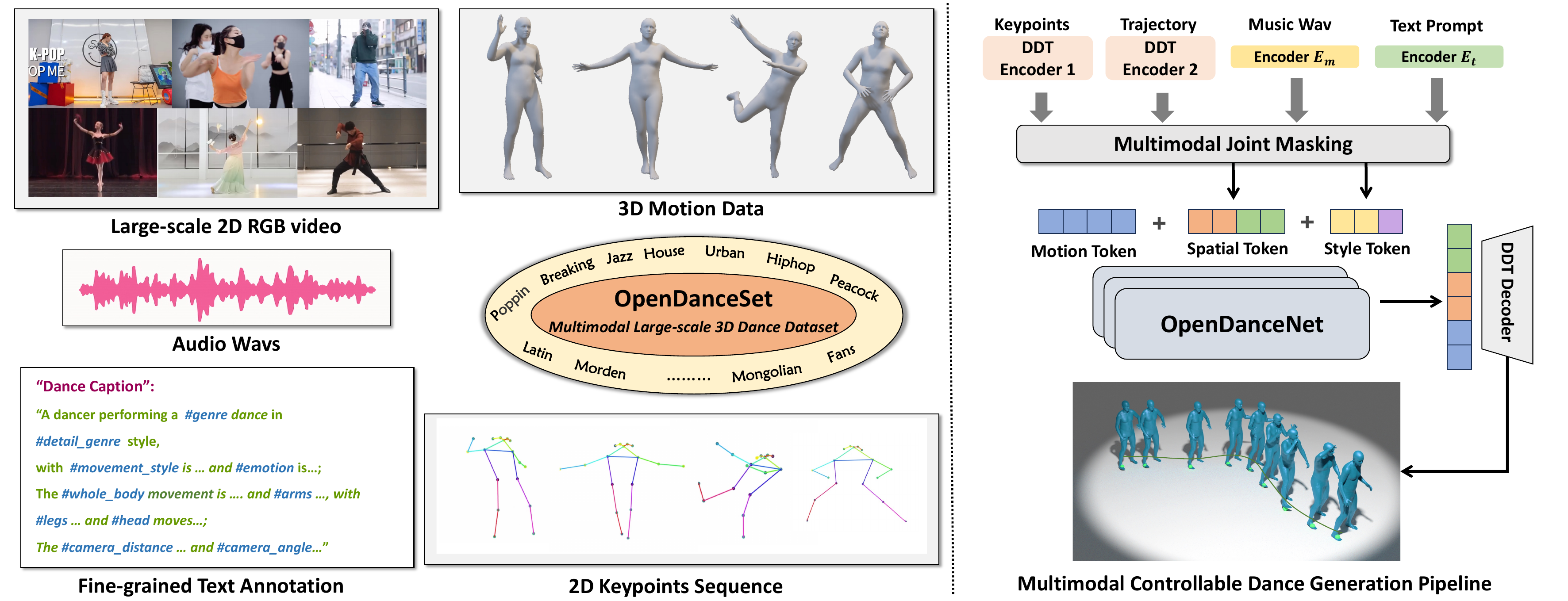}
    \vspace{-10pt}
    \captionof{figure}{We present a multimodal large-scale human dance dataset \textit{\dataset} and develop the masked modeling framework \textit{\model} for controllable and flexible dance generation conditioned on any "Music+X" setting (X: 2D keypoints, trajectory, and texts).}
    \label{fig:teaser}
\end{center}
}]
{
  \renewcommand{\thefootnote}{\fnsymbol{footnote}}
  \footnotetext[1]{Equal contribution} 
}
\begin{abstract}
Music-driven 3D dance generation offers significant creative potential, yet practical applications demand versatile and multimodal control. As the highly dynamic and complex human motion covering various styles and genres, dance generation requires satisfying diverse conditions beyond just music (\eg, spatial trajectories, keyframe gestures, or style descriptions). However, the absence of a large-scale and richly annotated dataset severely hinders progress. In this paper, we build \textit{\dataset}, an extensive human dance dataset comprising over 100 hours across 14 genres and 147 subjects. Each sample has rich annotations to facilitate robust cross-modal learning: 3D motion, paired music, 2D keypoints, trajectories, and expert-annotated text descriptions. Furthermore, we propose \textit{\model}, a unified masked modeling framework for controllable dance generation, including a disentangled auto-encoder and a multimodal joint-prediction Transformer. \model supports generation conditioned on music and arbitrary combinations of text, keypoints, or trajectories. Comprehensive experiments demonstrate that our work achieves high-fidelity synthesis with strong diversity and realistic physical contacts, while also offering flexible control over spatial and stylistic conditions. Project Page: \url{https://open-dance.github.io}.
\end{abstract}

\vspace{-10pt}
\section{Introduction}
\label{sec:intro}

Music-driven 3D dance generation shows immense potential for empowering applications such as virtual avatars in video games, immersive AR/VR experiences, and expressive digital humans.
Traditionally, producing such content relied on labor-intensive manual animation by 3D artists or professional dancers with motion capture systems, both of which are costly and difficult to scale.

Recent works~\cite{EDGEEditableDance2022Tseng,AIChoreographerMusic2021Li,Bailando3DDance2022Siyao,TM2DBimodalityDriven2023Gong,sun2020deepdance,ChoreoGraphMusicconditionedAutomatic2022Au,ChoreoMasterChoreographyorientedMusicdriven2021Chen,ListenDenoiseAction2023Alexanderson,PCDancePosturecontrollableMusicdriven2022Gao,MusicDrivenGroupChoreography2023Le,RhythmDancerMusicDriven2021Aristidou,DanceEditorIterative2025zhang,AlignYour2025fan} have made significant progress by using diffusion or autoregressive models and corresponding music-dance paired datasets~\cite{AIChoreographerMusic2021Li,FineDanceFinegrainedChoreography2023Li,EnchantDanceUnveilingPotential2023Han,sun2020deepdance,DanceMelodyLSTMautoencoder2018Tang,alemi2017groovenet,ChoreoMasterChoreographyorientedMusicdriven2021Chen,SoulDance2025li} to synthesize dance motion without much handcrafts.
Despite this promise, these deep generative models largely fail to provide the flexible controllability essential for practical use. This limitation stems from two fundamental bottlenecks:

The scarcity of richly-annotated and large-scale multimodal data. Most existing datasets~\cite{FineDanceFinegrainedChoreography2023Li,zhuang2022music2dance,DanceMelodyLSTMautoencoder2018Tang,alemi2017groovenet,SoulDance2025li} are collected from dancers using motion capture system, resulting in limited data scale and motion diversity for each genre (most are less than 1 hour for each genre). Furthermore, the lack of paired multimodal annotations (\eg, text, 2D keypoints) in these datasets makes diverse conditional generation impossible. While some works~\cite{MotionMixWeaklySupervisedDiffusion2024Hoang,UDEUnifiedDriving2022Zhou,TM2DBimodalityDriven2023Gong} attempt to bypass this by mixing text-to-motion datasets~\cite{kit-ml,humanml3d} during training, this is a suboptimal solution because of the different distributions of normal human motion data and diverse dance data.

The lack of a multimodal controllable generative model. During choreography, professional artists often need precise spatial control (\eg, key action, stage positioning) and style control (\eg, music beats, genre, movement styles). Therefore, a better AI choreography model is supposed to support flexible conditions. However, the majority of current generative models are music-only, without precise user control. While some recent works introduce limited capabilities like spatial editing~\cite{EDGEEditableDance2022Tseng, pinyoanuntapong2024mmm, guo2024momask} or seed-motion refinement~\cite{AIChoreographerMusic2021Li}, they lack a unified framework that can handle arbitrary combinations of diverse, multimodal conditions (\eg, text, keyframes, and trajectories). 
This leads to uncontrollable and low-utility results that cannot be aligned with specific user preferences.

\begin{table*}[t]
\centering
\caption{Comparison of previous 3D dance datasets. Time/Genre is the mean time (hours) for each genre. Our \dataset has more detailed and comprehensive annotations, more dancers with diverse movements, and longer duration.}
\vspace{-10pt}
\label{tab:dance_datasets}
\resizebox{0.7\linewidth}{!}{
\begin{tabular}{@{}lcccccccccc@{}}
\toprule
\textbf{Dataset} & \textbf{Subject} & \textbf{Genre} & \textbf{3D Pos/Rot} & \textbf{2D Kpt} & \textbf{Text} & \textbf{Time/Genre} & \textbf{Hours}\\ 
\midrule
GrooveNet~\cite{alemi2017groovenet} & 1 & 1 
& \cmark /\xmark & \xmark &\xmark & 0.38 & 0.38 \\
D. w. Melody~\cite{DanceMelodyLSTMautoencoder2018Tang} & - & 5 
& \cmark /\xmark & \xmark & \xmark & 0.40 & 1.56 \\
EA-MUD~\cite{eamud} & - & 4 
& \cmark /\cmark & \xmark & \xmark & 0.08 & 0.35 \\
DanceNet~\cite{zhuang2022music2dance} & 2 & 2 
& \cmark /\xmark & \xmark & \xmark & 0.48 & 0.96 \\
MMD~\cite{ChoreoMasterChoreographyorientedMusicdriven2021Chen} & - & 4 
& \cmark /\cmark & \xmark & \xmark & 2.47 & 9.91 \\ 
PhantomDance~\cite{DanceFormerMusicConditioned2022Li} & - & 13 
& \cmark /\xmark & \xmark & \xmark & 0.74 & 9.6 \\
AIST++ \cite{AIChoreographerMusic2021Li} & 30 & 10 
& \cmark /\cmark & \cmark & \xmark & 0.52 & 5.2 \\
FineDance~\cite{FineDanceFinegrainedChoreography2023Li} & {27} & \textbf{22} 
& \cmark /\cmark & \xmark & \xmark & {0.54} & {14.6} \\
PopDanceSet~\cite{luo2024popdg} & 132 & {19} 
& \cmark /\cmark & \cmark & \xmark & 0.18 & 3.56 \\
SoulDance~\cite{SoulDance2025li}    & - & {15} 
& \cmark /\cmark & \cmark & \xmark & 0.83 & 12.5 \\
\midrule
\textbf{\dataset (Ours)}  & \textbf{147} & {14} 
& \cmark /\cmark & \cmark & \cmark & \textbf{7.16} & \textbf{100.26} \\
\bottomrule\end{tabular}}
\vspace{-10pt}
\end{table*}

To address the data scarcity, we first build a large-scale and richly annotated multimodal dance dataset \textbf{\dataset}. As summarized in \Cref{tab:dance_datasets}, \dataset comprises 100.26 hours of 3D dance sequences from collected RGB videos, including 14 detailed genres. Unlike prior datasets~\cite{FineDanceFinegrainedChoreography2023Li,SoulDance2025li,ChoreoMasterChoreographyorientedMusicdriven2021Chen} that rely on costly MoCap systems, our approach (enabled by recent progress in video-based motion datasets~\cite{MotionxLargescale2023lin,MotionxLargescale2025zhang}) allows for significantly greater scale and diversity from immense online videos.
Crucially, it provides a comprehensive set of paired annotations essential for controllable generation: 3D motion, paired music, 2D keypoints, 3D trajectories, and textual descriptions, including the gender, genre, environment, and action style annotations from professional artists, human annotators, and VLM. These annotations cover both \textit{stylistic signals} (music, text) and \textit{spatial signals} (keypoints, trajectories), forming a diverse dataset for the task.

Second, to leverage the proposed data and enable flexible controllability under arbitrary multimodal conditions, we propose a unified masked modeling model \textbf{\model}. 
We apply music and text annotation as style control signals, and 2D keypoints sequence and 3D trajectory as spatial control signals.
We design a Disentangled Dance Tokenizer (DDT) to quantize motion and spatial signals into latent tokens, facilitating better alignment between motion token and spatial signal tokens. Then we utilize a Multi-Condition Transformer~(MCT) to learn the complex interplay between stylistic conditions and spatial conditions in latent space. To alleviate imbalanced multi-modal learning during MCT training, we propose the masked joint prediction mechanism to jointly predict motion tokens and spatial signal tokens, which facilitates better modeling performance. An efficient iterative per-step refinement is applied to further improve physical plausibility and control performance.
This architecture allows accepting music and arbitrary combinations of control signals (text, keypoints, or trajectories) and generates high-fidelity dance even from sparse or incomplete inputs.

In summary, our contributions can be summarized as:
\begin{itemize}
    \item We build a large-scale multimodal 3D dance dataset \dataset to improve the 3D dance generation controllability and diversity.
    \item \model is introduced to enable flexible generation from arbitrary combinations of multimodal conditions.
    \item Comprehensive experiments demonstrate the strong and competitive performance of our method in both generation quality and controllability.
\end{itemize}

\section{Related Works}
\label{sec:related_short}

\subsection{3D Music-driven Choreography Dataset}

Early MoCap datasets~\cite{alemi2017groovenet,DanceMelodyLSTMautoencoder2018Tang,zhuang2022music2dance} provide high-fidelity dance motion but offer limited style diversity and short durations. FineDance~\cite{FineDanceFinegrainedChoreography2023Li} and SoulDance~\cite{SoulDance2025li} extend coverage with professional dancers, yet remain under 15 hours due to lab-controlled capture. Video-based datasets scale more easily with improved pose estimation~\cite{xu2022vitpose,HRNet2019,HRNet2021PAMI,SMPLerXScalingExpressive2023Cai,shin2023wham}: AIST++~\cite{AIChoreographerMusic2021Li} reconstructs 5.2 hours from AIST~\cite{aist-dance-db} but has limited style variety; EA-MUD~\cite{eamud}, PhantomDance~\cite{DanceFormerMusicConditioned2022Li}, and MMD~\cite{ChoreoGraphMusicconditionedAutomatic2022Au} also remain moderate in scale and genres. EnchantDance~\cite{EnchantDanceUnveilingPotential2023Han} introduces a larger 70.32-hour SMPL~\cite{SMPLSkinnedMultiperson2015Loper} dataset but with only four genres, while PopDanceSet~\cite{luo2024popdg} increases subject diversity yet offers only 0.18 hour per genre. DanceEditor~\cite{DanceEditorIterative2025zhang} solely targets text-controlled editing, and other works~\cite{GroupDancerMusicMultiPeople2022Wang,DuolandoFollowerGPT2024Siyao,MusicDrivenGroupChoreography2023Le} focus on group choreography.

In contrast, \dataset provides large-scale, richly annotated 3D dance data with both stylistic (music, text) and spatial (keypoints, trajectory) signals, enabling diverse multimodal control for dance generation.

\subsection{Dance Generation Methods}
Music-to-motion generation is typically formulated as supervised cross-modal sequence modeling. Early works rely on recurrent and Transformer architectures: Tang \etal~\cite{DanceMelodyLSTMautoencoder2018Tang} regress acoustic features to 3D motions, AI Choreographer~\cite{AIChoreographerMusic2021Li} introduces a full-attention cross-modal Transformer, while GroupDancer~\cite{GroupDancerMusicMultiPeople2022Wang}, AIOZ-GDancer~\cite{MusicDrivenGroupChoreography2023Le}, and FreeDance~\cite{FreeDanceHarmonic2025zhao} extend to multi-dancer settings and global-local structures. 

From a generative perspective, Transflower~\cite{TransflowerProbabilisticAutoregressive2021Valle} uses normalizing flows, Bailando~\cite{Bailando3DDance2022Siyao} combines VQ-VAE codebooks with a GPT-based decoder, and ChoreoMaster~\cite{ChoreoMasterChoreographyorientedMusicdriven2021Chen}, PC-Dance~\cite{PCDancePosturecontrollableMusicdriven2022Gao}, and ChoreoGraph~\cite{ChoreoGraphMusicconditionedAutomatic2022Au} incorporate shared embeddings, posture control, and graph-based structure. More recently, diffusion-based models such as EDGE~\cite{EDGEEditableDance2022Tseng}, EnchantDance~\cite{EnchantDanceUnveilingPotential2023Han}, Controllable U-Net~\cite{guo2025controllable}, FineDance~\cite{FineDanceFinegrainedChoreography2023Li}, LODGE~\cite{li2024lodge}, and DanceMosaic~\cite{shah2025dancemosaic}, as well as SSM-based methods such as Danceba~\cite{AlignYour2025fan}, have enabled editable and long-horizon choreography.

In contrast to these methods, our framework \model unifies music, fine-grained text, 2D keypoints, and spatial trajectories into a single multimodal masked modeling paradigm, enabling precise and diverse controllable dance generation.

\subsection{Multimodal Dance Generation}
Beyond music-only conditioning, recent works explore multimodal inputs for more expressive and controllable dance generation. TM2D~\cite{TM2DBimodalityDriven2023Gong} fuses music and text via cross-attention; LM2D~\cite{yin2024lm2d} models lyrics--music alignment with dual-stream attention; LMM~\cite{zhang2024large} employs a transformer--diffusion backbone as a general motion foundation model. UDE~\cite{UDEUnifiedDriving2022Zhou,zhou2023unified} and MotionAnything~\cite{zhang2025motionanything} further unify diverse signals into shared latent spaces for conditional motion synthesis.
However, these approaches typically rely on hybrid training or latent fusion over limited multimodal dance data, providing only coarse control (e.g., style or category) rather than precise, frame-level guidance. 

The \model introduces a disentangled motion representation and multimodal joint prediction framework, combined with iterative refinement during inference, enabling flexible multimodal control.

\section{\dataset Dataset}
\label{sec:dataset}

\begin{figure*}[t]
    \centering
    \includegraphics[width=0.99\linewidth]{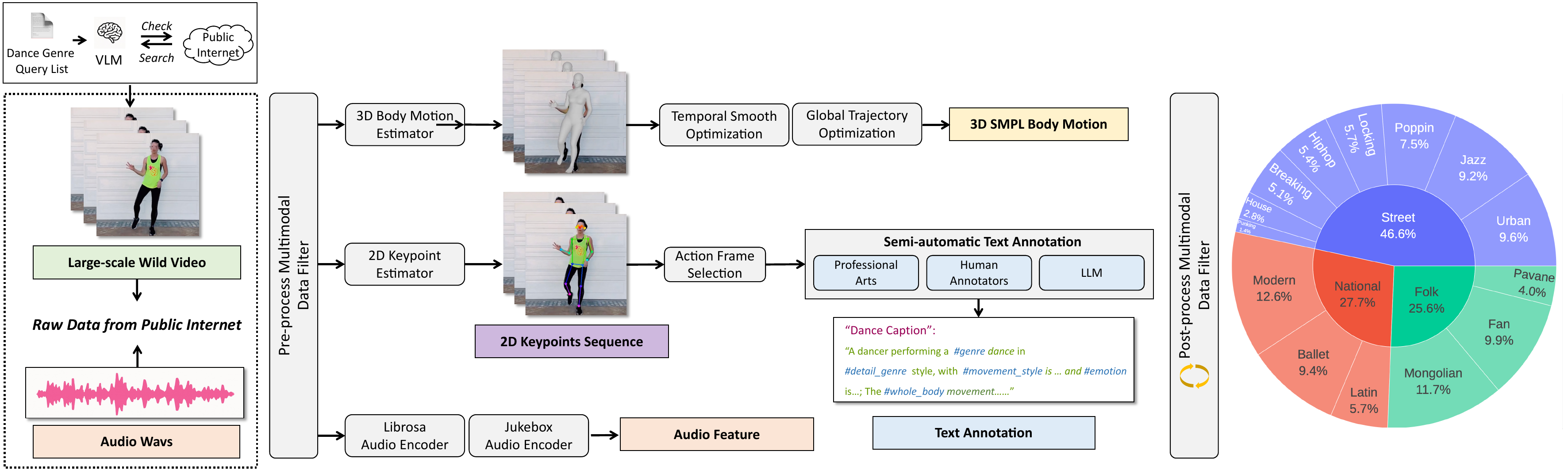}
    \vspace{-5pt}
    \caption{(Left) Overview of \dataset. We design an annotation pipeline to build the large-scale multimodal dance dataset, including 100.26 hours across 14 dance genres, with diverse annotations. (Right) The genre and detailed genre distribution in \dataset. }
    \vspace{-10pt}
    \label{fig:dataset}
\end{figure*}

\subsection{Overview}
As shown in Table~\ref{tab:dance_datasets}, we collect the multimodal dataset \dataset, which includes: RGB videos, audio, 2D keypoints, 3D SMPL motions, and text annotations. It comprises 100.26 hours, extracted from over 600 hours of wild videos, featuring 147 different dancers, and 41K long-duration (over 60 seconds) dance sequences. All annotations have been verified by professional artists to ensure quality, and all videos are resampled to 30 FPS.
With diverse annotations, \dataset can support other generative tasks, \eg, music-driven dance generation, text-driven dance generation, and pose-conditioned video generation.

We emphasize that all data was sourced from publicly available internet videos. 
Our released \dataset only contains derived 2D/3D representations, which do not include private, identifying information such as facial features or body shapes. 
To ensure reproducibility, we will release the original web links to the source videos. 
Users of these links are responsible for adhering to the original platforms' terms of service and privacy regulations. Therefore, \dataset does not involve any issues of privacy.


\subsection{Data Collection and Pre-process Filter}
Using a predefined list of dance genres and music style prompts provided by artists, we first employ the language model GPT to generate queries and check video captions for searching appropriate videos.
Upon acquiring approximately 600 hours of internet videos, in the pre-process multimodal data filter, we assess video-prompt alignment by human verification to filter out non-compliant videos. And we utilize the YOLOX \cite{ge2021yolox} to extract human bounding boxes, eliminating videos with incomplete bodies to isolate videos featuring solo dancers.

\subsection{Data Annotation Pipeline}
In our method, as shown in Figure~\ref{fig:dataset}, to efficiently capture an extensive range of dance motions from a vast array of videos, which is hard for human annotator, we introduce a semi-automatic annotation pipeline. This pipeline synergizes the strengths of a pre-trained estimator~\cite{ge2021yolox,xu2022vitpose,gvhmr2024shen}, LLM~\cite{Qwen2.5-VL}, human annotators, and professional artists to achieve high-quality whole-body motion capture, and the integration facilitates the annotation of 2D keypoints, 3D motion, audio features, and text descriptions.

\noindent \textbf{2D Keypoints.} 
Directly employing the 2D pose estimator~\cite{xu2022vitpose} to extract COCO keypoints from filtered videos may cause jittery problems due to the large variance of data. To overcome this issue, we apply a frame-level filtering to remove jitter poses, retaining clean videos that are suitable for capturing human motion.

\noindent \textbf{3D Motion Annotation.}
Based on the extracted keypoints, we estimate 3D dance movements using the SMPL~\cite{loperSMPLSkinnedMultiperson2015} model, a differentiable function that maps pose parameters $\boldsymbol{\theta}$, shape parameters $\boldsymbol{\beta}$, and root translation $\boldsymbol{\phi} \in \mathbb{R}^{1 \times 3}$ to a set of 3D human body mesh vertices $\mathbf{V} \in \mathbb{R}^{6890 \times 3}$ and 3D joint positions $\mathbf{X} \in \mathbb{R}^{J \times 3}$, where $J$ denotes the number of body joints.
Unlike the prior approach~ \cite{MusicDrivenGroupChoreography2023Le} that utilized an optimization-based approach~\cite{pavlakos2019expressive,SMPLX2019Pavlakos} for capturing robust motion parameters from single image, we do not follow these methods due to their slow frame-aware optimization process in massive videos. Furthermore, it is hard to provide world-grounded trajectories.
To mitigate these issues, we adopt the more efficient and accurate world-grounded 3D motion estimator~\cite{gvhmr2024shen}, a learning-based 3D pose estimator, to obtain more pixel-aligned and robust motion parameters (local joint parameters $\{\boldsymbol{\theta}\}_{t=1}^T$ and global translation $\{\boldsymbol{\phi_t}\}_{t=1}^T$, and beta parameters $\boldsymbol{\beta}$ for each sequence). 

\noindent \textbf{Text Descriptions.}
To obtain detailed text descriptions for each motion sequence, we allocate the annotation tasks based on the annotators levels of expertise and the difficulty of the task. 
Specifically, we employ: (i). Professional artists are responsible for annotating the three main genres and 14 detailed genres from a predefined list with identified genre\_id and d\_genre\_id.
(ii). Human annotators without professional dance knowledge are required to describe the start and end times of the dance video, gender, and the movement style for each video.
(iii). LLM~\cite{Qwen2.5-VL} is utilized to generate detailed descriptions of the body movements, including arms, legs, head and whole-body movements, as well as the dance style. The LLM annotation is queried by COCO keypoints visualization frames.

\noindent \textbf{Audio Features.}
We employ Jukebox~\cite{jukebox} and Librosa~\cite{librosa} to extract audio features from WAV files, where Jukebox can produce 4800 high-dimensional features to support better parsing of music. Librosa can extract beats and accompanying 35 low-dimensional features, which can support fast inference.



\subsection{Post-optimization and Dataset Refinement}
To further remove low-quality motion annotations, we design a post-processing and filtering pipeline.
For post-optimization, we first suppress temporal jitter by applying a Kalman filter, which improves the temporal smoothness of the captured motions.
Besides, we employ the Physical Foot Contact (PFC) score~\cite{EDGEEditableDance2022Tseng} as a penalty term during the optimization process. This further refines the motion by detecting when the foot contacts the ground, which helps prevent the foot-skating artifact and makes motions such as walking or running more realistic.

\begin{figure}[t]
    \centering
    \includegraphics[width=\linewidth]{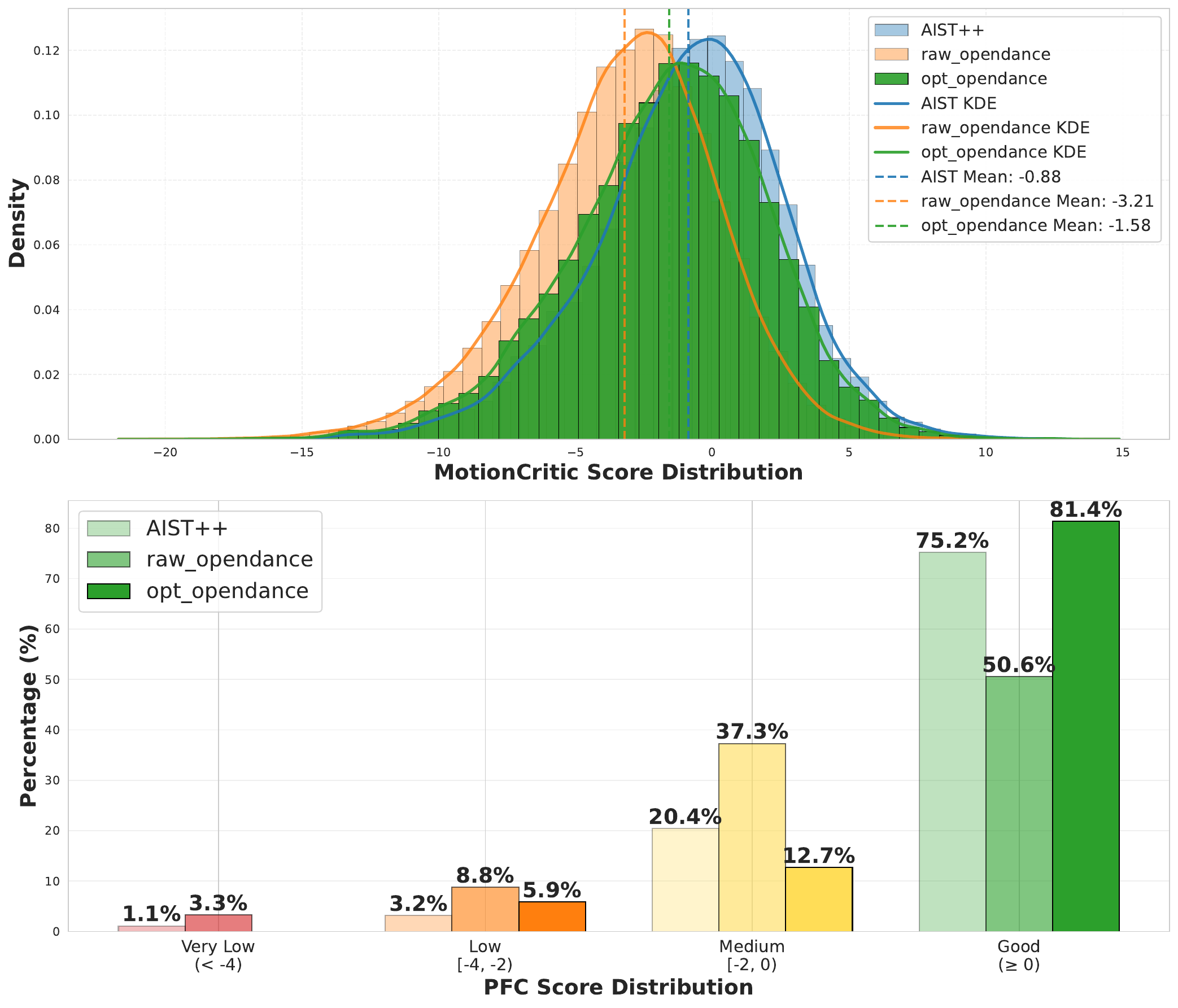}
    \vspace{-10pt}
    \caption{Comparison of processed \dataset, before filtered \dataset, and AIST++. After post-optimization and filtering, \dataset fits dance data distribution and achieves better physical performance.}
    \label{fig:data_process}
    \vspace{-15pt}
\end{figure}

For extracted data filtering, we first cut the first and last seconds of each clip based on human-annotated start and end times, eliminating initial artifacts (e.g., jitter) from the dancer's entry.
We then filter the processed data based on several metrics: a jitter score, a stillness score, the PFC score, and a human-alignment score.
Low-quality samples are removed by applying a predefined threshold to each score.
For the human-alignment score, we employ the MotionCritic~\cite{motioncritic2025} metric.
However, this critic model was trained on text-to-motion data, which has a significant distribution gap compared to our music-to-dance data (as shown in \Cref{fig:data_process}).
Therefore, to create a reliable filter, we use the high-quality AIST++ dataset~\cite{AIChoreographerMusic2021Li} as a reference distribution.
We first compute the MotionCritic score distribution on AIST++ and then filter our dataset to approximate this target distribution, primarily by removing low-scoring samples.

During training, we slice the data to facilitate efficient training processes. To ensure coherence between music and motion, we require the shortest motion sequence must exceed 5 seconds. Furthermore, we employ data augmentation by using larger stride steps to sample genres with abundant dance data, such as \textit{Street}, and dense steps for genres with scarce data, to ensure a balanced representation across different genres. 

\begin{figure*}[t!]
    \centering
    \includegraphics[width=1.0\linewidth]{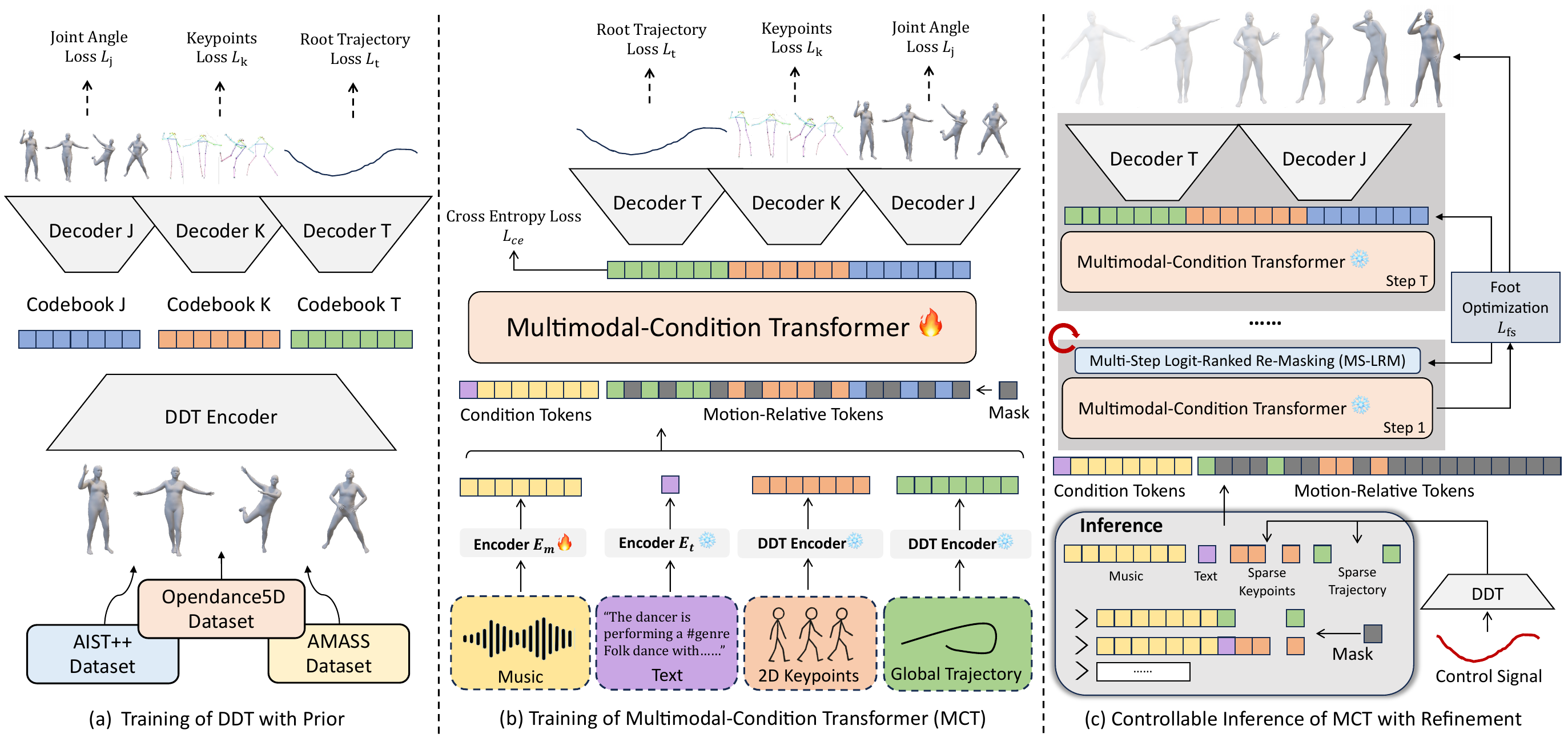}
    \vspace{-10pt}
    \caption{Overview of \model, a masked-modeling-based dance generation framework.
    (a) We first train a Disentangled Dance Tokenizer (DDT) to quantize spatial signals (joint rotations, global trajectories, and 2D keypoints) into discrete tokens.
    (b) Then the Multimodal-Condition Transformer (MCT) is trained by randomly sampling subsets of control modalities and applying token-level masks over trajectories, 2D keypoints, and motion tokens, enabling the model to handle diverse condition combinations while generating coherent dance motions.
    (c) At inference time, \model supports arbitrary configurations of input conditions for flexible multimodal control, while Multi-Step Logit-Ranked Re-Masking (MS-LRM) and footstep optimization progressively refine the generated motions and improve their physical plausibility.}
    \vspace{-10pt}
    \label{fig:model}
\end{figure*}

\section{\model}
\label{sec:method}

Prior dance generation methods are predominantly trained on single-modality datasets such as AIST++~\cite{AIChoreographerMusic2021Li}, and therefore cannot support controllable generation conditioned on specified body movements at particular timestamps, the dancer’s position on stage at specific moments, or language descriptions. This limitation substantially reduces their practicality. Furthermore, we observe that different control modalities—such as language, motion, and spatial position—contribute to supervisory signals of unequal strength, and naïvely optimizing them jointly fails to capture the precise relationships between these conditions and the resulting motions. 
In this section, we introduce \textbf{\model}, a controllable dance generation framework designed to address these issues. \model first disentangles and models multiple modalities separately and then unifies them through a masked prediction paradigm, yielding state-of-the-art controllable dance generation performance. Specifically, \Cref{sec:dvq} presents the training of our \textbf{Disentangled Dance Tokenizer (DDT)}, and \Cref{sec:transformer} details the architecture of the \textbf{Multimodal-Condition Transformer (MCT)}. \Cref{sec:supervision} then describes how we leverage spatial supervision to strengthen alignment among different modalities, and \Cref{sec:inference} explains how we employ \textbf{Multi-Step Logit-Ranked Re-Masking (MS-LRM)} together with a per-step refinement strategy during inference to further enhance the quality and controllability of generated dances.

\subsection{Disentangled Dance Tokenizer (DDT)}
\label{sec:dvq}
Pretraining the DDT is crucial for controllable dance generation. Existing methods are fundamentally limited by (i) insufficient motion diversity---most clips depict near-static, repetitive movements, often leading to out-of-distribution failures under controllable settings; and (ii) weak alignment between continuous control signals (e.g., trajectories, keypoints) and the latent spaces used by generative models. DDT is specifically designed to address both issues.

We first unify \dataset, AIST++, and the more diverse AMASS dataset~\cite{mahmood2019amass} under a shared motion representation and train a single DDT on this combined corpus. The model takes three modalities as input: joint rotations, global trajectory, and COCO-style 2D keypoints~\cite{coco2014lin}. For a motion sequence of length $T$ with joint rotations $J \in \mathbb{R}^{T \times D_j}$, keypoints $K \in \mathbb{R}^{T \times D_k}$, and global trajectory $X \in \mathbb{R}^{T \times D_x}$, the encoder maps each modality independently to latent features $z_i \in \mathbb{R}^{T' \times d}$ for $i \in \{J, K, X\}$, which are then quantized into discrete tokens $\hat{z}_i$ using three learned codebooks $C_i = \{c_n\}_{n=1}^N$. These modality-specific tokens can be stacked into a unified discrete representation $\hat{z} \in \mathbb{R}^{3 \times T' \times d}$.

Crucially, we avoid early cross-modal fusion in the encoder, enabling sparse, frame-level constraints (e.g., partial 2D keypoints or global trajectories) to be directly padded and encoded as tokens through their respective DDT branches. This disentangled design provides an explicit and consistent mapping from each control modality to its token sequence, forming a robust foundation for precise and flexible multi-conditional dance generation.

\subsection{Multimodal Masked Joint Prediction}
\label{sec:transformer}

Previous generative masked motion models only predict motion tokens. In contrast, our Multimodal-Condition Transformer (MCT) is designed as a joint predictor. Because the dependencies between the disentangled token streams are highly structured, we empirically find that simply treating 2D keypoints and global trajectory as additional conditions—on par with music and text—and only generating motion tokens, as in previous work, is insufficient for the model to learn from them effectively. We argue that this is due to the fact that 2D keypoints and global trajectory impose strict frame-level constraints on motion and require the model to synthesize accurate 3D position. This means the neural network prefers to learn the generation process from coarse, high-level conditions (e.g., stylistic signals) and ignores more challenging, fine-grained conditions (e.g., spatial signals). 

To address this issue, MCT is trained not only to generate human motion from a masked motion sequence, but also to reconstruct the ground-truth tokens of 2D keypoints and global trajectory from their masked versions. At the same time, to prevent the motion stream from overly depending on any single modality, we apply random modality-level masking during training with probability $p_{\text{mask}}$.

Given an input sequence of length $T$, the conditioning modalities are music $M \in \mathbb{R}^{T \times D_m}$, 2D keypoints $K \in \mathbb{R}^{T \times D_k}$, text $u \in \mathbb{R}^{D_{\text{text}}}$, and global trajectory $X \in \mathbb{R}^{T \times D_x}$. After tokenization, we obtain token streams $Z_{\text{music}}, Z_{\text{text}}, Z_{\text{traj}}, Z_{\text{kpts}}$ and concatenate them into
\begin{equation}
Z = [Z_{\text{music}}, Z_{\text{text}}, Z_{\text{traj}}, Z_{\text{kpts}}].
\end{equation}
The masked input sequence is then constructed as:
\begin{equation} 
\begin{aligned} 
Z_{\text{mask}} = \big[ & (Z_{\text{music}})_{p_{\text{mask}}},\; (Z_{\text{text}})_{p_{\text{mask}}}, \\ & \text{Mask}(Z_{\text{traj}}),\; \text{Mask}(Z_{\text{kpts}}) \big],
\end{aligned} 
\end{equation}
where $(\cdot)_{p_{\text{mask}}}$ denotes stochastic modality-level masking with probability $p_{\text{mask}}$, and $\mathrm{Mask}(\cdot)$ applies token-level masking for reconstruction. Here $Z$ is the full input sequence to MCT and $Z_{\text{mask}}$ is the corresponding sequence with special $[\text{MASK}]$ tokens.

Each modality is converted into tokens by a dedicated tokenizer: we use a Jukebox-style audio encoder for music, a CLIP-based encoder for text, and DDT encoders for 2D keypoints and global trajectory. During training, $[\text{MASK}]$ tokens are randomly inserted into the input token streams. MCT takes all (possibly masked) modalities as input, passes them through several self-attention Transformer layers, and jointly recovers the original dance motion, trajectory, and keypoint tokens.
\begin{equation}
\mathcal{L}_{\text{CE}}^{\text{mask}}
= - \mathbb{E}_{Z} \sum_{i \in \mathcal{M}} \log p_{\theta}(z_i \mid Z_{\text{mask}}).
\label{eq:csloss}
\end{equation}
Let $\mathcal{M}$ denote the set of masked positions in $Z$. After predicting the likelihoods of all masked token IDs, the model is optimized with a cross-entropy loss in \Cref{eq:csloss}.

\subsection{Spatial Auxiliary Supervision}
\label{sec:supervision}
To fully exploit the masked modeling scheme of MCT, we introduce spatial supervision and physics-based constraints to tightly couple global trajectory, 2D keypoints, and joint rotations, thereby reducing foot sliding and improving realism. During training, we use Gumbel-Softmax~\cite{jang2017categorical} to obtain differentiable motion token samples, decode them with DDT into predicted trajectories $X_{\text{pred}}$ and 2D keypoints $K_{\text{pred}}$, and supervise them against ground truth. We further enforce 3D consistency via a forward kinematics loss and a foot contact loss that penalizes inconsistent foot behavior.

Formally, the objective includes the following.
\begin{align}
\mathcal{L}_{\text{traj}} &= \lambda_{g} \left\lVert X - X_{\text{pred}} \right\rVert_1, \\
\mathcal{L}_{\text{kpts}} &= \lambda_{\text{kpts}} \left\lVert K - K_{\text{pred}} \right\rVert_1, \\
\mathcal{L}_{\text{fk}} &= \lambda_{\text{fk}} \left\lVert F(J, X) - F(J_{\text{pred}}, X_{\text{pred}}) \right\rVert_1, \\
\mathcal{L}_{\text{con}} &= \frac{\lambda_{\text{con}}}{N} \sum_{i=1}^{N}
\Big\lVert \big(F(J, X) - F(J_{\text{pred}}, X_{\text{pred}})\big)\cdot b^i \Big\rVert
\end{align}
where $F(\cdot)$ maps joint rotations and trajectories to 3D joint positions, $b^i$ denotes predefined binary foot-contact labels, and $N$ is the sequence length. The weighted sum of these terms enforces accurate reconstruction, coherent kinematics, and contact-consistent dance motions.

\subsection{Inference with Per-Step Refinement}
\label{sec:inference}

Since we apply multimodal masking during training, MCT naturally supports flexible inference under arbitrary combinations of conditions and control signals. Users can provide sparse frame-level constraints (e.g., partial 2D keypoints or trajectories), which are tokenized by DDT and injected into MCT as hard conditions.

At inference time, we perform iterative masked prediction with $N$ refinement steps. In step $i$, MCT predicts distributions for all $[\text{MASK}]$ positions given the current token sequence. We maintain per-token confidence and apply \emph{Multi-Step Logit-Ranked Re-Masking} (MS-LRM) to re-mask low-confidence tokens across all iterations, leading to more stable and accurate results than methods that only re-mask tokens from the latest step (e.g., MoMask~\cite{guo2024momask}).

To mitigate foot sliding arising from decoupled rotations and trajectories, each step includes a physics-aware refinement: we sample motion tokens via Gumbel-Softmax, decode them with DDT, compute 3D joints via FK, and update logits using the foot sliding loss $\mathcal{L}_{\text{fs}}$:
\begin{equation}
\hat{\mathbf{e}}_{\text{logits}} = \mathbf{e}_{\text{logits}} - \alpha \,\nabla_{\mathbf{e}_{\text{logits}}} \mathcal{L}_{\text{fs}}.
\end{equation}
followed by resampling. In the final step, we further refine the motion embedding:
\begin{equation}
\hat{\mathbf{e}}_{\text{motion}} = \mathbf{e}_{\text{motion}} - \alpha \,\nabla_{\mathbf{e}_{\text{motion}}} \mathcal{L}_{\text{fs}}.
\end{equation}
Finally, a lightweight post-processing module is applied to suppress residual foot-skating artifacts, yielding smooth, realistic motion with stable contacts.

\section{Experiments}
\label{sec:exp}

\begin{figure}[t]
    \centering
    \includegraphics[width=\linewidth]{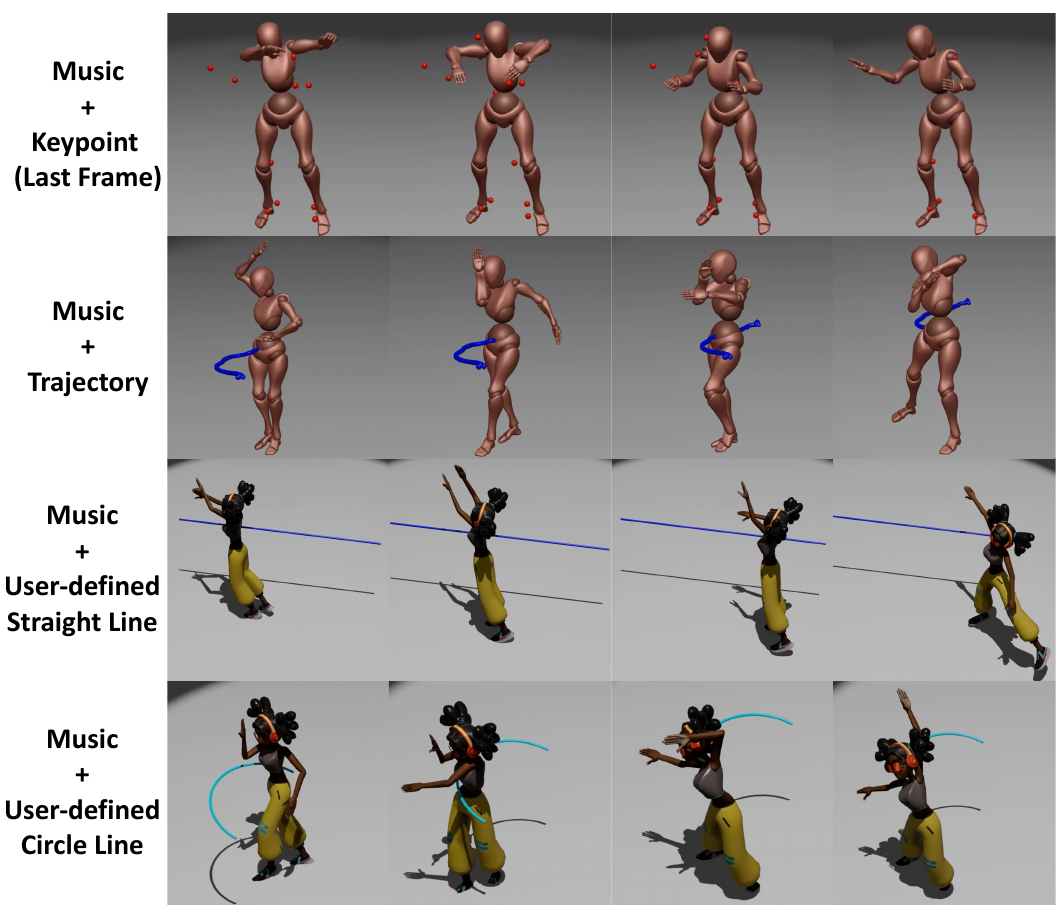}
    \vspace{-10pt}
    \caption{Different spatial control signals visualization. The top two rows use GT keypoint (only last frame) and GT trajectory as conditions, and bottom two rows use random geometry trajectory. We recommend viewing videos in supplementary.}
    \label{fig:vis_control}
    \vspace{-15pt}
\end{figure}

\subsection{Implementation Details}
During training, we apply data augmentation with non-uniform sampling strides: larger strides for genres with abundant data (e.g., \textit{Street}) and denser sampling for underrepresented genres. We evaluate our method on both AIST++~\cite{AIChoreographerMusic2021Li} and \dataset: following the official splits, we preprocess AIST++ into 5-second clips at 30 FPS and \dataset into 10-second clips at 30 FPS. Poses are represented with a 24-joint SMPL model using 6-DoF rotations per joint~\cite{zhou2019continuity}, a 3D root translation, and binary heel/toe contact labels as in EDGE~\cite{EDGEEditableDance2022Tseng}, yielding per-frame vectors $x \in \mathbb{R}^{24 \times 6 + 4 + 3}$. Music is encoded with Jukebox~\cite{jukebox} and text with CLIP~\cite{clip}. Our model is trained on a single NVIDIA RTX 4090 GPU (24 GB).

For evaluation, we report $\mathrm{FID}_k$ and $\mathrm{FID}_g$ following~\cite{AIChoreographerMusic2021Li}, $\mathrm{Div}_k$ and $\mathrm{Div}_g$ as in Bailando~\cite{Bailando3DDance2022Siyao}, and Beat Alignment Score (BAS) and Physical Foot Contact (PFC)~\cite{EDGEEditableDance2022Tseng} to assess rhythmic alignment and physical plausibility. Details are provided in the Appendix.

\subsection{Comparisons}
We conduct comparative experiments of several existing
state-of-the-art methods, including FACT~\cite{AIChoreographerMusic2021Li}, Bailando~\cite{Bailando3DDance2022Siyao},
EDGE~\cite{EDGEEditableDance2022Tseng}, LODGE~\cite{li2024lodge}, FineDance~\cite{FineDanceFinegrainedChoreography2023Li}, MoMask~\cite{guo2024momask}, TM2D~\cite{TM2DBimodalityDriven2023Gong} on the AIST++ and our \dataset datasets.


On AIST++~\cite{AIChoreographerMusic2021Li} (Table~\ref{tab:comp_aistpp}), \model attains the lowest PFC, the best $\mathrm{FID}_k = 24.82$, competitive $\mathrm{FID}_g = 12.54$, and the highest BAS (0.2513), indicating strong physical plausibility and rhythm alignment. On \dataset (Table~\ref{tab:comp_opendance}), it further achieves the best FID$_k$ (23.19), the best FID$_m$ (7.72) after optimization, strong diversity (Div$_k$ = 7.82, Div$_g$ = 6.41), and the highest BAS (0.2472), outperforming EDGE~\cite{EDGEEditableDance2022Tseng}, TM2D~\cite{TM2DBimodalityDriven2023Gong}, and other baselines. These results together show that \model provides the best overall trade-off in realism, fidelity, diversity, and beat alignment across both datasets.

\begin{table}[t]
    \centering
    \caption{Comparison on AIST++ dataset~\cite{AIChoreographerMusic2021Li}.}
    \vspace{-8pt}
    \renewcommand\arraystretch{1.0}
    \resizebox{\linewidth}{!}{
    \begin{tabular}{l|cccccc}
    \toprule
    \textbf{Method} & \textbf{PFC $\downarrow$} & \textbf{FID$_k$ $\downarrow$} & \textbf{FID$_g$ $\downarrow$} & \textbf{Div$_k$ $\rightarrow$} & \textbf{Div$_g$ $\rightarrow$} & \textbf{BAS $\uparrow$} \\ 
    \hline 
    Ground Truth & 1.332 & 17.10 & 10.60 & 8.19 & 7.45 & 0.2374 \\
    \hline 
    FACT~\cite{AIChoreographerMusic2021Li} & 2.254 & 35.35 & 22.11 & 5.94 & 6.18 & 0.2209 \\
    Bailando~\cite{Bailando3DDance2022Siyao} & 1.754 & 28.16 & \textbf{9.62} & \textbf{7.83} & 6.34 & 0.2332 \\
    EDGE~\cite{EDGEEditableDance2022Tseng} & 1.536 & 31.82 & 22.16 & 8.73 & \textbf{7.18} & 0.2043 \\
    LODGE~\cite{li2024lodge} & -  & 37.09	& 18.79	& 5.58	& 4.85  & 0.2423	\\
    MoMask~\cite{guo2024momask} & 1.648 & 44.92	& 26.20	& 4.67	& 1.06	& 0.2312 \\
    \midrule
    \model(Ours) & \textbf{1.140} & \textbf{24.82} & 12.54 & 5.24 & 5.29 & \textbf{0.2513} \\
    \bottomrule
    \end{tabular}
    }
    \label{tab:comp_aistpp}
    \vspace{-10pt}
\end{table}

\begin{table}[t]
    \centering
    \caption{Comparison on our proposed \dataset Dataset.}
    \vspace{-8pt}
    \label{tab:comp_opendance}
    \renewcommand\arraystretch{1.0}
    \resizebox{\linewidth}{!}{
    \begin{tabular}{lccccccc}
    \hline
    \textbf{Method}       & \textbf{PFC $\downarrow$} & \textbf{FID$_k$ $\downarrow$} & \textbf{FID$_g$ $\downarrow$} & \textbf{Div$_k$ $\rightarrow$} & \textbf{Div$_g$ $\rightarrow$} & \textbf{BAS $\uparrow$} \\ \hline
    Ground Truth          & 0.1578 & 8.05 & 2.98 & 11.49 & 7.82 & 0.2453 \\
    \hline 
    EDGE~\cite{EDGEEditableDance2022Tseng} & \textbf{0.2386} & 36.42 & 9.97	& 7.21	& \textbf{7.29}	& 0.2372 \\
    FineDance~\cite{FineDanceFinegrainedChoreography2023Li} & 0.3366 & 26.69 & 9.98 & 7.50 & 6.13 & 0.2387 \\
    MoMask~\cite{guo2024momask} & 0.3281	& 61.11	& 20.19	& 2.03	& 1.27	& 0.2344 \\
    TM2D~\cite{TM2DBimodalityDriven2023Gong} & 2.8794	& 69.95	& 23.42	& 3.53	& 1.27	& 0.2201 \\
    \hline
    \model(Ours) & 0.3462 & \textbf{23.19}	& 11.89	& \textbf{7.82}	& 6.41	& \textbf{0.2472}\\
    Ours (+ Refinement) & 0.2733 & 37.40	& \textbf{7.72}	& 6.72	& 6.78	& 0.2389\\
    \bottomrule
    \end{tabular}
    }
    \vspace{-15pt}
\end{table}

\subsection{Ablation Study}

\textbf{Ablation on effectiveness of inference control.}
We examine the impact of adding different spatial control signals during inference on \dataset. As shown in \Cref{tab:control_signal_type}, incorporating spatial control into \model provides clear guidance for the generated dances while preserving overall motion quality. These results indicate that \model not only handles multimodal inputs, but also effectively leverages spatial control for fine-grained guidance.

\textbf{Ablation on joint prediction mechanism of MCT.}
We further analyze the role of joint modality training in MCT. As reported in \Cref{tab:ablation_joint_predict}, removing joint prediction of trajectory and keypoints together with motion substantially degrades generation quality. Joint prediction of spatial signals and motion tokens enables MCT to better capture cross-modal relationships, mitigate foot-sliding artifacts, and strengthen its overall generative capability.

\textbf{Ablation on multiple-condition training.}
We conduct an ablation study on \dataset to assess the effect of training with multiple conditions. We compare models trained with a single condition (e.g., music only) against those trained with different condition combinations, as summarized in \Cref{tab:ablation_multi_cond}. The results show that multi-condition training significantly improves the ability to generate diverse, high-quality dance motions, as reflected by better FID and diversity scores.

\textbf{Ablation on loss functions for Transformer training.}
We use the music-conditioned \model trained with cross-entropy loss alone on \dataset as the baseline to evaluate each additional supervision term (see \Cref{tab:ablation_loss_trans}). The results demonstrate that these auxiliary losses provide consistent guidance for MCT training, enabling the model to better fit the data.

\begin{table}[t]
    \centering
    \caption{Effect of control signal types on AIST and \dataset. Ablation results are reported on a subset of \dataset.}
    \vspace{-8pt}
    \label{tab:control_signal_type}
    \resizebox{\linewidth}{!}{
    \begin{tabular}{c c|cccccc}
    \toprule
    \textbf{Dataset} & \textbf{Control} 
    & \textbf{PFC $\downarrow$} 
    & \textbf{FID$_k$ $\downarrow$} 
    & \textbf{FID$_g$ $\downarrow$} 
    & \textbf{BAS $\uparrow$}
    & \textbf{Traj dist. $\downarrow$}
    & \textbf{Kpts dist. $\downarrow$} \\
    \midrule
    \multirow{2}{*}{AIST}
        & None        & 1.264 & 23.36 & 11.62 & 0.2224 & 0.6508 & --     \\
        & Traj        & 1.421 & 42.52 & 16.88 & 0.2208 & 0.0334 & --     \\
    \midrule
    \multirow{4}{*}{\dataset}
        & None        & 0.179 & 48.02 & 15.24 & 0.2402 & 0.1545 & 0.1400 \\
        & Traj        & 0.197 & 49.95 & 14.50 & 0.2381 & 0.0141 & 0.1176 \\
        & Kpts        & 0.172 & 44.28 & 13.04 & 0.2385 & 0.1216 & 0.0444 \\
        & Traj + Kpts & 0.181 & 48.03 & 11.99 & 0.2339 & 0.0141 & 0.0444 \\
    \bottomrule
    \end{tabular}
    }
\end{table}

\begin{table}[t]
    \centering
    \caption{Ablation on joint prediction mechanism. Traj indicates global trajectory condition.}
    \vspace{-8pt}
    \label{tab:ablation_joint_predict}
    \resizebox{\linewidth}{!}{
    \begin{tabular}{ccc|cccccc}
    \toprule
    \textbf{Music} & \textbf{Traj} & \textbf{Kpts2D} & \textbf{PFC $\downarrow$} & \textbf{FID$_k$ $\downarrow$} & \textbf{FID$_g$ $\downarrow$} 
    & \textbf{Div$_k$ $\rightarrow$} & \textbf{Div$_g$ $\rightarrow$} & \textbf{BAS $\uparrow$}   \\
    \midrule
    \checkmark &  & 
      & 0.6919 & 171.69 & 39.11 & 8.45 & 7.38 & 0.2417 \\
    \checkmark & \checkmark &
      & 0.3053 & 51.69 & 23.83 & 5.54 & 8.12 & 0.2253  \\
    \checkmark & \checkmark & \checkmark
      & 0.3498 & 47.18 & 21.34 & 6.18 & 7.91 & 0.2345  \\
    \bottomrule
    \end{tabular} 
    }
    \vspace{-5pt}
\end{table}

\begin{table}[t]
    \centering
    \caption{Ablation on multi-condition training.}
    \vspace{-8pt}
    \label{tab:ablation_multi_cond}
    \resizebox{\linewidth}{!}{
    \begin{tabular}{cccc|ccccccccc}
    \toprule
    \textbf{Music} & \textbf{Traj} & \textbf{Kpts2D} & \textbf{Text} 
    & \textbf{PFC $\downarrow$} & \textbf{FID$_k$ $\downarrow$} & \textbf{FID$_g$ $\downarrow$} 
    & \textbf{Div$_k$ $\rightarrow$} & \textbf{Div$_g$ $\rightarrow$} & \textbf{BAS $\uparrow$}   \\
    \midrule
    \checkmark &  &  &  
      & 0.3584 & 102.87 & 37.42 & 11.27 & 10.30 & 0.2335  \\
    \checkmark & \checkmark &  &  
      & 0.2457 & 52.54 & 28.76 & 5.86 & 9.30 & 0.2191  \\
    \checkmark & \checkmark & \checkmark & 
      & 0.3049 & 49.31 & 21.09 & 5.77 & 7.86 & 0.2182  \\
    \checkmark & \checkmark & \checkmark & \checkmark 
      & 0.3202 & 48.46 & 21.79 & 5.95 & 8.00 & 0.2288  \\
    \bottomrule
    \end{tabular} 
    }
    \vspace{-5pt}
\end{table}

\begin{table}[t]
    \centering
    \caption{Ablation on loss functions on Transformer training.}
    \vspace{-8pt}
    \label{tab:ablation_loss_trans}
    \resizebox{\linewidth}{!}{
    \begin{tabular}{cccc|cccccc}
    \toprule
    $L_{con}$ & $L_{fk}$    & $L_{traj}$ & $L_{kpts}$ 
                       & \textbf{PFC} $\downarrow$   & \textbf{FID$_k$} $\downarrow$ & \textbf{FID$_g$} $\downarrow$ 
                       & \textbf{BAS} $\uparrow$ \\
    \midrule
     &  &  &  
      & 0.3142 & 48.91 & 22.32 & 0.2191  \\
    \checkmark &  &  &  
      & 0.3046 & 47.92 & 21.90 & 0.2180  \\
    \checkmark & \checkmark &  & 
      & 0.3039 & 47.83 & 21.80 & 0.2198  \\
    \checkmark & \checkmark & \checkmark &
      & 0.3083 & 48.39 & 21.58 & 0.2220  \\
    \checkmark & \checkmark & \checkmark & \checkmark 
     & 0.2966 & 46.92 & 20.38 & 0.2194  \\
    \bottomrule
    \end{tabular}
    }
    \vspace{-10pt}
\end{table}

\section{Limitation and Future Work}
\label{sec:limitation}
Despite the advances demonstrated, our work opens several avenues for future research. First, \dataset does not capture detailed hand–finger articulations or facial expressions. In addition, although we collect raw RGB videos as one modality, we have not yet benchmarked or adapted state-of-the-art video generation models on \dataset due to our current research focus and safety constraints. Finally, we currently encode all textual annotations into a single CLIP~\cite{clip} token to obtain latent condition, disregarding distinctions between genre and body-movement descriptions. Future work can explore vision–language architectures that has better tokenizer to support more detailed textual descriptions, improving editing ability of texts.

\section{Conclusion}
We present OpenDance, a multimodal-condition dance generation framework that introduces a novel multimodal dataset \dataset with 100.26 hours of in-the-wild recordings across 14 genres and 5 synchronized modalities, together with a unified masked modeling architecture \model for flexible and controllable conditional generation. Extensive experiments show that our method produces diverse, realistic dance motions that closely follow user-defined conditions, outperforming existing approaches in both fidelity and controllability.

\newpage
{
    \small
    \bibliographystyle{ieeenat_fullname}
    \bibliography{main}
}


\end{document}